\documentclass{article} 
\usepackage{iclr2021_conference,times}

\usepackage{microtype}
\usepackage{graphicx}
\usepackage{natbib}
\usepackage{booktabs} 
\usepackage{multirow}       
\usepackage{makecell}       
\usepackage{wrapfig} 
\usepackage{xcolor}
\usepackage{subcaption}
\usepackage{algorithm}
\usepackage{algorithmic}

\usepackage{amsmath,amsfonts,bm}

\def\gB{{\mathcal{B}}}

\def\gD{{\mathcal{D}}}

\def\gL{{\mathcal{L}}}

\def\gO{{\mathcal{O}}}
\def\gP{{\mathcal{P}}}

\def\gW{{\mathcal{W}}}

\def\sC{{\mathbb{C}}}

\def\0{\mathbf{0}}
\def\1{\mathbf{1}}

\def\Figref#1{Fig.~\ref{#1}}
\def\Secref#1{Sec.~\ref{#1}}
\def\Algref#1{Alg.~\ref{#1}}
\def\Tabref#1{Table~\ref{#1}}
\def\Eqref#1{Eq.~\eqref{#1}}
\def\NAME{{AutoHAS}}

\newcommand{\TT}[1]{\texttt{#1}}
\newcommand{\BF}[1]{\textbf{#1}}

\newcommand{\BLUE}[1]{\textcolor{blue}{#1}}

\def\BLUE{\textcolor{blue}}
\DeclareFontFamily{U}{mathx}{\hyphenchar\font45}
\DeclareFontShape{U}{mathx}{m}{n}{<-> mathx10}{}
\DeclareSymbolFont{mathx}{U}{mathx}{m}{n}
\DeclareMathAccent{\widebar}{0}{mathx}{"73}

\usepackage{hyperref}

\usepackage{url}

\title{AutoHAS: Efficient Hyperparameter and Architecture Search}

\author{Xuanyi Dong$^{\dagger\ddagger}$\thanks{Work done as a research intern at Google.}, Mingxing Tan$^{\dagger}$, Adams Wei Yu$^{\dagger}$, Daiyi Peng$^{\dagger}$, Bogdan Gabrys$^{\ddagger}$, Quoc Le$^{\dagger}$\\
$^{\dagger}$ Google Research, Brain Team, $^{\ddagger}$ AAI, University of Technology Sydney
}

\iclrfinalcopy 
\begin{document}

\maketitle

\begin{abstract}
Efficient hyperparameter or architecture search methods have shown remarkable results, but each of them is only applicable to searching for either hyperparameters (HPs) or architectures. 
In this work, we propose a unified pipeline, AutoHAS, to efficiently search for both architectures and hyperparameters.
AutoHAS learns to alternately update the shared network weights and a reinforcement learning (RL) controller, which learns the probability distribution for the architecture candidates and HP candidates. A temporary weight is introduced to store the updated weight from the selected HPs (by the controller), and a validation accuracy based on this temporary weight serves as a reward to update the controller.
In experiments, we show AutoHAS is efficient and generalizable to different search spaces, baselines and datasets.
In particular, AutoHAS can improve the accuracy over popular network architectures, such as ResNet and EfficientNet, on CIFAR-10/100, ImageNet, and four more other datasets.
\end{abstract}

\section{Introduction}\label{sec:intro}

Deep learning models require intensive efforts in optimizing architectures and hyperparameters. Standard hyperparameter optimization methods, such as grid search, random search or Bayesian optimization, are inefficient because they are \textit{multi-trial}: different configurations are tried in parallel to find the best configuration. As these methods are expensive, there is a trend towards more efficient, single-trial methods for specific hyperparameters. For example, the learning rate can be optimized with the hypergradient method~\citep{baydin2018hypergradient}.
Similarly, many architecture search methods started out multi-trial~\citep{zoph2017neural,baker2017designing,real2019regularized}, but more recent proposals are single-trial~\citep{pham2018efficient,liu2019darts}.  
These efficient methods, however, sacrifice generality: each method only works for one aspect or a subset of the hyperparameters or architectures.

In this paper, we generalize those efficient, single-trial methods to include both architectures and a broader range of hyperparameters\footnote{In this paper, hyperparameters refer all design choices that will affect the training procedure of a model, such as learning rate, weight decay, optimizer, dropout, augmentation policy, etc.}. 
One important benefit of the generalization is that we can have a general, efficient method for hyperparameter optimization as a special case. 
Another benefit is that we can now search for both hyperparameters and architectures in a single model. Practically, this means that our method is an improvement over neural architecture search (NAS) because each model can potentially be coupled with its own best hyperparameters, thus achieving comparable or even better performance than existing NAS with \textit{fixed} hyperparameters.  


\begin{table*}[t!]
\centering
\small
\setlength{\tabcolsep}{2.8pt}
\begin{tabular}{l c c c c c c}
\toprule
        & learning rate (LR) & weight decay & augmentation & dropout & architecture & \textbf{efficient} \\
\midrule
Bayesian & $\surd$ & $\surd$ & $\surd$ & $\surd$ & $\surd$ & $\times$ \\
RL or Evolution & $\surd$ & $\surd$ & $\surd$ & $\surd$ & $\surd$ & $\times$ \\
PBT & $\surd$ & $\surd$ & $\surd$ & $\surd$ & $\times$ & $\times$ \\
Gradient Descent on LR & $\surd$ & $\times$  & $\times$ & $\times$ & $\times$ & $\surd$ \\
Hypergradient & $\times$ & $\surd$  & $\surd$ & $\times$ & $\surd$ & $\surd$ \\
NAS (Weight Sharing) & $\times$ & $\times$ & $\times$ & $\times$ & $\surd$ & $\surd$ \\
\midrule
  \textbf{\NAME} & $\surd$ & $\surd$ & $\surd$ & $\surd$ & $\surd$ & $\surd$ \\
\bottomrule
\end{tabular}
\vspace{-2mm}
\caption{
We compare the different aspects of each algorithm.
We consider a search algorithm is efficient if it can complete the search within less than $10\times$ computational costs than training a single model.
Bayesian optimization, traditional RL and evolutionary algorithms are applicable to all kinds of hyperparameters and architectures, but they are computationally expensive.
Population based training (PBT) methods~\citep{jaderberg2017population,li2019generalized} utilized the incomplete observations to accelerate the evolutionary algorithms for HPO, but they are still far from efficient.
Recent hypergradient-based HPO methods~\citep{lorraine2020optimizing,pedregosa2016hyperparameter,shaban2019truncated,baydin2018hypergradient} and weight sharing-based NAS methods~\citep{liu2019darts,xie2019snas,pham2018efficient,dong2019search} are efficient but sacrifice the generality.
Our {\NAME} takes care of both efficiency and generality.
}
\vspace{-3mm}
\label{table:compare-method-attribute}
\end{table*}

\begin{figure*}[t!]
\center
\includegraphics[width=\linewidth]{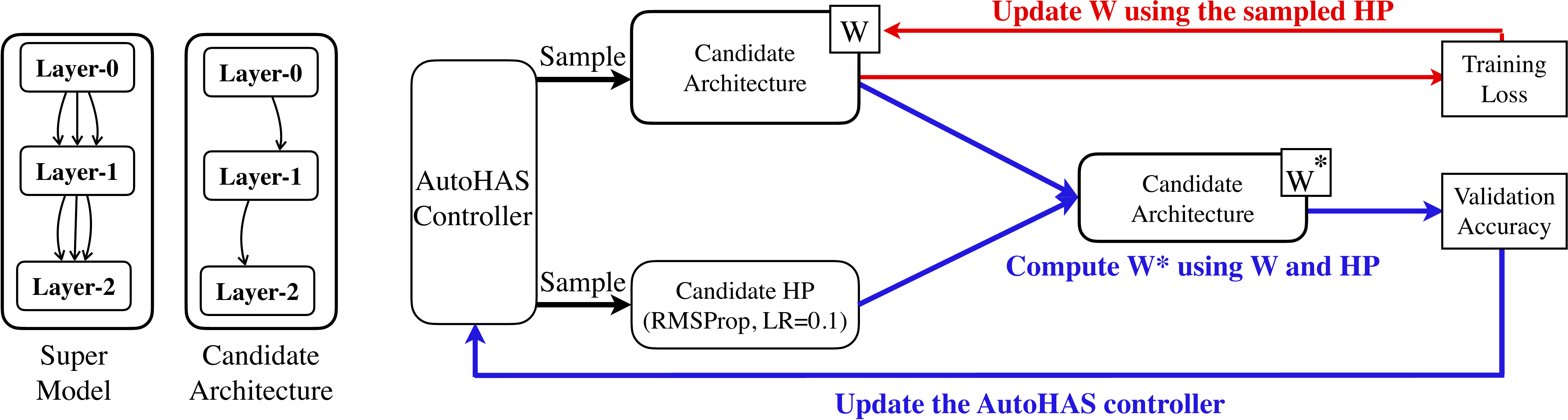}
\caption{
\BF{The overview of {\NAME}.}
LEFT: Each candidate architecture's weights are shared with a super model, where each candidate is a sub model within this super model.
RIGHT: 
During the search, AutoHAS alternates between optimizing the shared weights of super model $\gW$ and updating the controller.
It also creates temporary weights $\gW^{*}$ by optimizing the sampled candidate architecture using the sampled candidate hyperparameter (HP). This $\gW^{*}$ will be used to compute the validation accuracy as a reward so as to update the AutoHAS controller to select better candidates. Finally, $\gW^{*}$ is discarded after updating the controller so as not to affect the original $\gW$.
}
\label{fig:autohas-high-level}
\end{figure*}

To this end, we propose {\NAME}, an efficient hyperparameter and architecture search framework.
It is, to the best of our knowledge, \textit{the first method that can efficiently handle architecture space, hyperparameter space, or the joint search space.} {\NAME} utilizes the weight sharing technique proposed by~\cite{pham2018efficient}.
The main idea is to train a super model, where each candidate in the architecture space is its sub-model.
Using a super model can avoid training millions of candidates from scratch~\citep{liu2019darts,pham2018efficient}.
{\NAME} extends its scope from architecture search to both architecture and hyperparameter search.
We not only share the weights of super model with each architecture but also share this super model across hyperparameters.
At each search step, {\NAME} optimizes the sampled sub-model by a combination of the sampled hyperparameter choices, and the shared weights of super model serves as a good initialization for all hyperparameters at the next step of search (see \Figref{fig:autohas-high-level} and \Secref{sec:autohas}).
In order to decouple the shared network weights ($\gW$ in \Figref{fig:autohas-high-level}) and controller optimization, we also propose to create temporary weights for evaluating the sampled hyperparameters and updating the controller.

A challenge here is that architecture choices (e.g. kernel size) are often categorical values whereas hyperparameter choices (e.g. learning rate) are often continuous values. 
To address the mixture of categorical and continuous search spaces, we first discretize the continuous hyperparameters into a linear combination of multiple categorical basis. The discretization allows us to unify architecture and hyperparameter choices during search.
As explained below, we will use a reinforcement learning (RL) method to search over these discretized choices in \Figref{fig:autohas-high-level}.
The probability distribution over all candidates is naturally learnt by the RL controller, and it is used as the coefficient in the linear combination to find the best architecture and hyperparameters.

{\NAME} shows generalizability, efficiency and scaling to large datasets in experiments. It consistently improves the baselines' accuracy when searching for architectures, hyperparameters, and both on CIFAR-10, CIFAR-100, ImageNet, Place365, etc.
In experiments, {\NAME} shows non-trivial improvements on \textbf{eight} datasets, such as 0.8\% accuracy gain on highly-optimized EfficientNet and 11\% accuracy gain on less-optimized models.
{\NAME} reduces the search cost by over $10\times$ than random search and Bayesian search.
We also summarize the benefits of our {\NAME} over other search methods in \Tabref{table:compare-method-attribute}.

\section{{\NAME}}\label{sec:autohas}



{\NAME} can handle the general case of NAS and HPO -- able to find both architecture $\alpha$ and hyperparameters $h$ that achieve high performance on the validation set $\gD_{val}$.
This objective can be formulated as a bi-level optimization problem:
\begin{align}\label{eq:autohas}
 \min_{\alpha, h} \gL(\alpha, \omega_{\alpha,h}^{*}, \gD_{val}) ~\hspace{1mm}~
    \mathrm{s.t.} \hspace{1mm}\omega_{\alpha,h}^{*} = f_{h}(\alpha, \gD_{train}) ,
\end{align}
\noindent where $\gL$ is the objective function (e.g., cross-entropy loss).
$\gD_{train}$ and $\gD_{val}$ denote the training data and the validation data, respectively.
$f_{h}$ represents the algorithm with hyperparameters $h$ to obtain the optimal weights $\omega_{\alpha,h}^{*}$.
For example, $f_{h}$ could be using SGD to minimize the training loss, where $h$ denotes the hyperparameters of SGD.
In this case, $\omega_{\alpha,h}^{*}$ is the final optimized weights after the SGD converged.

{\NAME} generalizes both NAS and HPO by introducing a broader search space.
On the one hand, NAS is a special case of HAS, where $h$ is fixed in \Eqref{eq:autohas}.
On the other hand, HPO is a special case of HAS, where $\alpha$ is fixed in \Eqref{eq:autohas}.

\subsection{Unified Representation of Hyperparameters and Architectures}\label{sec:autohas-represent}

The search space in {\NAME} is a Cartesian product of the architecture and hyperparameter candidates.
To search over this mixed search space, we need a unified representation of different searchable components, i.e., architectures, learning rates, optimizer, etc.

\textbf{Architectures Search Space~~}
We use the simplest case as an example.
First of all, let the set of predefined candidate operations (e.g., 3x3 convolution, pooling, etc.) be $\gO=\{O_{1}, O_{2}, ..., O_{n}\}$, where the cardinality of $\gO$ is $n$ for each layer in the architecture.
Suppose an architecture is constructed by stacking multiple layers, each layer takes a tensor $F$ as input and output $\pi(F)$, which serves as the next layer's input.
$\pi \in \gO$ denotes the operation at a layer and might be different at different layers.
Then a candidate architecture $\alpha$ is essentially the sequence for all layers $\{\pi\}$.
Further, a layer can be represented as a linear combination of the operations in $\gO$ as follows:
\begin{align}\label{eq:nas-arch-space}
  \pi(F) = \sum\nolimits_{i=1}^{n} C^{\alpha}_{i}~O_{i}( F ) ~\hspace{1mm}~
    \mathrm{s.t.} \hspace{1mm} \sum\nolimits_{i=1}^{n} C^{\alpha}_{i}=1, C^{\alpha}_{i} \in \{0, 1\} ,
\end{align}
\noindent where $C^{\alpha}_{i}$ (the $i$-th element of the vector $C^{\alpha}$) is the coefficient of operation $O_i$ for a layer.

\textbf{Hyperparameter Search Space~~}
Now we can define the hyperparameter search space in a similar way.
The major difference is that we have to consider both categorical and continuous cases:
\begin{align}\label{eq:nas-hp-space}
    h = \sum\nolimits_{i=1}^{m} C^{h}_{i}~\gB_{i} ~\hspace{1mm}~
    \mathrm{s.t.} \hspace{1mm} \sum\nolimits_{i=1}^{m} C^{h}_{i}=1, ~\textrm{and}~C^{h}_{i}~\in\begin{cases}
     {[0, 1]}, & \textrm{if~continuous} \\
     {\{0, 1\}}, & \textrm{if~categorical}
    \end{cases},
\end{align}
\noindent where $\gB$ is a predefined set of hyperparameter basis with the cardinality of $m$ and $\gB_{i}$ is the $i$-th basis in $\gB$.
$C^{h}_{i}$ (the $i$-th element of the vector $C^{h}$) is the coefficient of hyperparameter basis $\gB_{i}$. If we have a continuous hyperparameter, we have to discretize it into a linear combination of basis and unify both categorical and continuous. For example, for weight decay, $\gB$ could be \{1e-1, 1e-2, 1e-3\}, and therefore, all possible weight decay values can be represented as a linear combination over $\gB$.
For categorical hyperparameters, taking the optimizer as an example, $\gB$ could be \{Adam, SGD, RMSProp\}.
In this case, a constraint on $C^{h}_{i}$ is applied: $C^{h}_{i} \in \{0, 1\}$ as in \Eqref{eq:nas-hp-space}.

\subsection{Efficient Hyperparameter and Architecture Search}\label{sec:autohas-algo}

Given the discretizing strategy in \Secref{sec:autohas-represent}, each candidate in the search space can be represented by the value of $\sC = \{C^{\alpha}~\textrm{for all layers}, C^{h}~\textrm{for all hyperparameters}\}$, which represents the coefficients for all architecture and hyperparameter choices.
As a result, AutoHAS converts the searching problem to obtaining the coefficients $\sC$.
Below we will introduce how to calculate these coefficients.

AutoHAS applies reinforcement learning together with weight sharing to search over the discretized space.
During search, we learn a controller to sample the candidate architecture and hyperparameters from the discretized space.
In AutoHAS, this controller is parameterized by a collection of independent multinomial variables $\gP=\{P^{\alpha}~\textrm{for all layers}, P^{h}~\textrm{for all hyperparameters}\}$
, which draws the probability distribution of the discretized space.
AutoHAS also leverages a super model to share weights $\gW$ among all candidate architectures, where each candidate is a sub-model in this super model.
Furthermore, AutoHAS extends the scope of weight sharing from architecture to hyperparameters, where $\gW$ also serves as the initialization for the algorithm $f_{h}$.

AutoHAS alternates between learning the shared weights $\gW$ and learning the controller using REINFORCE~\citep{williams1992simple}.
Specifically, at each iteration, the controller samples a candidate --- an architecture $\alpha$ and basis hyperparameter $h\in\gB$.
We estimate its quality $Q(\alpha, h)$ by utilizing the temporary weights $\gW^{*}_{\alpha}$, which are generated by applying the gradients from training loss to the original weights $\gW_{\alpha}$ of the architecture $\alpha$ with hyperparameters $h$.
This estimated quality is used as a reward to update the controller's parameters $\gP$ via REINFORCE.
Then, we optimize the shared weights $\gW$ by minimizing the training loss calculated by the sampled architecture.
More details can be found in \Algref{alg:autohas} in Appendix.

\textbf{Discussion~~}
In practice, the training of shared weights in efficient NAS often suffers from the instability problem~\citep{liu2019darts,dong2019search,zela2020understanding}, and that for HAS can be more pronounced.
To make AutoHAS more stable, we will sample tens of candidate pairs of $\alpha$ and $h$, and average the gradients from all these pairs for $\gW$ as well as that for $\gP$ to update $\gW$ and $\gP$.
In this way, the high variance of the gradients from different samples can be significantly reduced.
In experiments, this strategy is achieved via training on the distribution system, such that each core can individually sample a different pair.

During the aforementioned strategy, the temporary weights allows us to effectively decouple the shared network weights and controller optimization.
If we directly override $\gW$ without using temporary weights $\gW^{*}$, it will pollute the shared weights update.
This is because that different sampled pairs will use different weights for overriding and quarrel with each other.

\subsection{Deriving Hyperparameters and Architecture}\label{sec:autohas-derive}

After {\NAME} learns $\gP=\{P^{\alpha}, P^{h}\}$, we can derive the coefficient $\sC$ as follows:
\begin{align}\label{eq:autohas-derive}
\vspace{-2mm}
C^{\alpha} = \textrm{one\_hot}({\arg\max}_{i}~P^{\alpha}) ~\hspace{2mm}~
    C^{h} = \begin{cases}
     P^{h} & \textrm{if~continuous} \\
     \textrm{one\_hot}({\arg\max}_{i}~P^{h}) & \textrm{if~categorical}
    \end{cases} ,
\vspace{-2mm}
\end{align}
\noindent Together with \Eqref{eq:nas-arch-space} and \Eqref{eq:nas-hp-space}, we can derive the final architecture $\alpha$ and hyperparameters $h$.
Intuitively speaking, the selected operation in the final architecture has the highest probability over other candidates, and so does the categorical hyperparameter.
For the continuous hyperparameter, the final one is the weighted sum of the learnt probability $P^{h}$ with its basis $\gB$.

To evaluate whether the AutoHAS-discovered $\alpha$ and $h$ is good or not, we will use $h$ to re-train $\alpha$ on the whole training set and report its performance on the test sets.

\section{Experimental Analysis}

\begin{table}[b!]
\scriptsize
  \centering
  \vspace{-4mm}
  \setlength{\tabcolsep}{3.4pt}
  \begin{tabular}{l c c c c c c}
    \toprule
    & \multirow{2}{*}{\#Params (MB)} & \multirow{2}{*}{\#FLOPs (M)} & \multirow{2}{*}{\makecell{ImageNet\\Accuracy (\%)}} & \multicolumn{2}{c}{Search Cost}  \\
    &  & &  &  Memory (GB) & Time (TPU Hour) \\
    \midrule
 Baseline model & 1.5 & 35.9 & 50.96 &   1.0 (Training Cost) & 44.8 (Training Cost) \\
    \midrule
 AutoHAS (Differentiable)  & 1.5 & 36.1 & 52.17 &  6.1  & 92.8  \\
 AutoHAS (REINFORCE)       & 1.5 & 36.3 & 53.01 &  \textbf{1.8}  & \textbf{54.4}\\
 \bottomrule
  \end{tabular}
  \vspace{-2mm}
  \caption{
  AutoHAS Differentiable Search vs. AutoHAS REINFORCE Search on ImageNet.
  }
  \vspace{-2mm}
  \label{table:100-epochs-g-vs-r-main}
\end{table}


{\NAME} is a unified, efficient, and general framework for joint architecture and hyperparameter search.
We show its applicability and efficiency when integrated with RL-based and differentiable-based searching algorithm in \Tabref{table:100-epochs-g-vs-r-main}. It is also easy to switch from RL to a Bayesian-based searching algorithm, which will be investigated in the future.

In \Tabref{table:100-epochs-g-vs-r-main}, we use the popular MobileNetV2~\citep{sandler2018mobilenetv2} as our baseline model.
We compare its performance with AutoHAS-discovered architecture and hyperparameters.
On the same par of computational resources as training MobileNet-V2, AutoHAS finds a better model. In addition, AutoHAS (REINFORCE) is more efficient than AutoHAS (Differentiable).
More empirical analysis can be found in Appendix. 

\paragraph{Acknowledgements}
We want to thank Gabriel Bender, Hanxiao Liu, Hieu Pham, Ruoming Pang, Barret Zoph and Yanqi Zhou for their help and feedback.

\bibliography{abrv,ms}
\bibliographystyle{iclr2021_conference}

\appendix

\section{Explain AutoHAS Algorithm}

AutoHAS alternates between optimizing the super model (parameterized by the shared weights $\gW$) and the AutoHAS controller (parameterized by the multinomial variables $\gP$). Once AutoHAS completes the training of controller, we will follow \Secref{sec:autohas-derive} to derive the final architecture and hyperparameters.

To optimize the AutoHAS controller, it samples a candidate --- an architecture $\alpha$ and basis hyperparameter $h\in\gB$.
Taking the architecture $\alpha$'s weights $\gW_{\alpha}$ as the initialization, we apply hyperparameter $h$ to update $\gW_{\alpha}$ by one or multiple gradient descent steps to obtain the temporary weights $\gW^{*}_{\alpha}$.
We use the temporary weights $\gW^{*}_{\alpha}$ to calculate the quality $Q(\alpha, h)$ to update the AutoHAS controller.
After that, the temporary weights $\gW^{*}_{\alpha}$ will be discarded and not affect the original weights $\gW_{\alpha}$.
When we target better accuracy for HAS, the quality $Q(\alpha, h)$ is the accuracy of a batch of validation data.
When we target on FLOP-constrained problem, the quality $Q(\alpha, h)$ becomes a hybrid one combing both accuracy and FLOPs, such as the hard exponential reward function in MNasNet~\citep{tan2019mnasnet} or absolute reward function in TuNAS~\citep{bender2020can}.
In \Algref{alg:autohas}, we show an example of using the simplest RL algorithm -- REINFORCE -- to optimize the controller.
AutoHAS is general and can be integrated with a differentiable searching algorithm to optimize AutoHAS controller\footnote{See more technical details in our preliminary version: https://arxiv.org/pdf/2006.03656v1.pdf and empirical comparison in \Tabref{table:100-epochs-g-vs-r}.}.
Bayesian optimization can also be easily integrated into AutoHAS, where the acquisition function serves to sample candidates and the quality used to improve the surrogate model in Bayesian optimization. We would investigate these extensions in future.

To optimize the super model's weights, AutoHAS controller samples a candidate --- an architecture $\alpha$ and basis hyperparameter $h\in\gB$.
Then, AutoHAS use the optimization algorithm represented by $h$ to update the weights $\gW_{\alpha}$ in the super model corresponding to the architecture $\alpha$.

In practice, we average the gradients for multiple samples to update the super model or the controller. This can stabilize the training of AutoHAS and can be naturally implemented by using parallel computation of multiple GPU/TPU devices.

Note that, in the main paper, we sometimes omit ${\alpha}$ in $\gW^{*}_{\alpha}$ and $\gW_{\alpha}$ for simplicity.

\textbf{Scope of AutoHAS}: AutoHAS can search for most kinds of hyperparameters, including learning rate scheduler, momentum coefficient, weight decay, dropout ratio, data augmentation policy, etc.
However, it can not be used to search for weight initialization.

\textbf{Comparison with weight-sharing NAS}:
AutoHAS utilized the \textit{temporary weights} $\gW^{*}$ instead of the raw weights $\gW$ in \citep{pham2018efficient,li2020random,dong2019search,zela2020understanding} to compute the validation accuracy (loss for differentiable NAS).
This accuracy over $\gW^{*}$ evaluates ``the performance of both architecture and hyperparameters'' instead of ``the performance of only architecture'' in \citep{pham2018efficient,li2020random,dong2019search,zela2020understanding}.
In addition, AutoHAS generalizes both ENAS~\citep{pham2018efficient} and GDAS~\citep{dong2019search}.
ENAS can be viewed as a special case of AutoHAS that uses $\gW$ replacing $\gW^{*}$ and searches for architecture only.
GDAS can be viewed as a special case of AutoHAS (Differentiable Variant) that uses $\gW$ replacing $\gW^{*}$ and searches for architecture only.

\textbf{How to create the basis for hyperparameters?}
For the categorical hyperparameter, its basis is the set of candidate categorical values.
For the continuous hyperparameter, we typically uniform-choose 10 scalars around its default value, where the default value is the commonly used value in previous literature.

\begin{algorithm}[H]
\small
  \caption{The {\NAME} Algorithm.\\
  $\gW$ indicates the super model's weights.
  $\gW_{\alpha}$ indicates the weights of $\alpha$, which is a subset of $\gW$.
  }
  \label{alg:autohas}
  \begin{algorithmic}[1]
    \REQUIRE Split the available data into two disjoint sets: $\gD_{train}$ and $\gD_{val}$
    \STATE Randomly initialize the super model's weights $\gW$ and the controller's parameters $\gP$
    \WHILE{not converged}
    	  \STATE Sample ($\alpha, h\in\gB$) from the controller
    	  \STATE Compute the temporary weights $\gW^{*}_{\alpha}$ by applying $h$ on $\gW_{\alpha}$
    	  \STATE Calculate the quality $Q(\alpha, h)$ as the reward to update controller by REINFORCE
    	  \STATE \textit{[Optionally]} Re-sample ($\alpha, h\in\gB$) from the controller
    	  \STATE Optimize $\gW_{\alpha}$ by one step of gradient descent based on $f_{h}$ and $\gD_{train}$
        
    \ENDWHILE
    \STATE Derive the final architecture $\alpha$ and hyperparameters $h$ by $\gP$ (\Secref{sec:autohas-derive})
  \end{algorithmic}
\end{algorithm}

\section{Related Works}\label{sec:related}

We summarize the advantages of {\NAME} over other neural architecture search and hyperparameter search (aka., hyperparameter optimization) methods in \Tabref{table:compare-method-attribute}.

\textbf{\noindent Neural Architecture Search.} 
Since the seminal works~\citep{baker2017designing,zoph2017neural} show promising improvements over manually designed architectures, more efforts have been devoted to NAS.
The accuracy of NAS models has been improved by carefully designed search space~\citep{zoph2018learning}, better search method~\citep{real2019regularized}, or compound scaling~\citep{tan2019efficientnet}.
The model size and latency have been reduced by Pareto optimization~\citep{tan2019mnasnet,wu2019fbnet,cai2018proxylessnas,cai2020once} and enlarged search space of neural size~\citep{cai2020once}.
The efficiency of NAS algorithms has been improved by weight sharing~\citep{pham2018efficient}, differentiable optimization~\citep{liu2019darts}, or stochastic sampling~\citep{dong2019search,xie2019snas}.
As these NAS methods use fixed hyperparameters during search, we have empirically observed that they often lead to sub-optimal results, because different architectures tend to favor their own hyperparameters. 
In addition, even if the manual optimization of architecture design is avoided by NAS, they still need to tune the hyperparameters after a good architecture is discovered.

\textbf{\noindent Hyperparameter Optimization (HPO).}
Black-box and multi-fidelity HPO methods have a long standing history~\citep{bergstra2012random,hutter2009automated,hutter2011sequential,hutter2019automated,kohavi1995automatic}.
Black-box methods, e.g., grid search and random search~\citep{bergstra2012random}, regard the evaluation function as a black-box.
They sample some hyperparameters and evaluate them one by one to find the best.
Bayesian methods can make the sampling procedure in random search more efficient~\citep{jones1998efficient,shahriari2015taking,snoek2015scalable}.
They employ a surrogate model and an acquisition function to decide which candidate to evaluate next.
Multi-fidelity optimization methods accelerate the above methods by evaluating on a proxy task, e.g., using less training epochs or a subset of data~\citep{domhan2015speeding,jaderberg2017population,kohavi1995automatic,li2017hyperband}.
These HPO methods are computationally expensive to search for deep learning models.

Recently, gradient-based HPO methods have shown better efficiency~\citep{baydin2018hypergradient,lorraine2020optimizing}, by computing the gradient with respect to the hyperparameters.
For example, \cite{maclaurin2015gradient} calculate the extract gradients w.r.t. hyperparameters.
\cite{pedregosa2016hyperparameter} leverages the implicit function theorem to calculate approximate hypergradient. Following that, different approximation methods have been proposed~\citep{lorraine2020optimizing,pedregosa2016hyperparameter,shaban2019truncated}.
Despite of their efficiency, they can only be applied to differentiable hyperparameters such as weight decay, but not non-differentiable hyperparameters, such as learning rate~\citep{lorraine2020optimizing} or optimizer~\citep{shaban2019truncated}.
Our {\NAME} is not only as efficient as gradient-based HPO methods but also applicable to both differentiable and non-differentiable hyperparameters.
Moreover, we show significant improvements on state-of-the-art models with large-scale datasets, which supplements the lack of strong empirical evidence in previous HPO methods.

Our method is also related to population based training (PBT)~\citep{jaderberg2017population,li2019generalized}.
They are asynchronous optimization algorithms that jointly optimize
a population of models and their hyperparameters.
Our {\NAME} is more efficient and general than PBT methods due to two factors.
First, PBT needs to maintain a large number of model population in parallel, whereas {\NAME} only needs to handle a single model and its temporary weights at each iteration.
Second, PBT can not be directly used for architecture search.

Recent HPO methods also pay attention to the mixture search space~\citep{ru2020bayesian,daxberger20mixed}. However, due to their multi-trial nature, it is impractical to apply them for large-scale models and datasets.

\textbf{\noindent Hyperparameter and Architecture Search.}
Few approaches have been developed for the joint searching of hyperparameter and architecture~\citep{klein2019tabular,zela2018towards}.
However, they focus on small datasets and small search spaces. These methods are more computationally expensive than {\NAME}.
Concurrent to our {\NAME}, FBNet-V3~\citep{dai2020fbnetv3} learns an acquisition function to predict the performance for the pair of hyperparameter and architecture. They require to evaluate thousands of pairs to optimize this function and thus costs much more computational resources than ours.

\section{Experiments}\label{sec:exp-all}

We evaluate {\NAME} on eight datasets, including two large-scale datasets, ImageNet~\citep{deng2009imagenet} and Places365~\citep{zhou2017places}.
We will briefly introduce the experimental settings in \Secref{sec:autohas-exp-setting}.
We demonstrate the generalizability of {\NAME} in \Secref{sec:exp-gen} and show its efficiency in \Secref{sec:exp-efficient}.
Furthermore, we verify {\NAME}'s scalability in \Secref{sec:exp-practical}.
Lastly, we ablatively study {\NAME} in \Secref{sec:exp-abla}.

\subsection{Experimental Settings}\label{sec:autohas-exp-setting}



\begin{table*}[t!]
\scriptsize
\centering
\setlength{\tabcolsep}{4pt}
  \begin{tabular}{l l l l l l l}
    \toprule
 Method   & Search Space    & CIFAR-10  & CIFAR-100  & Stanford Cars & Oxford Flower & SUN-397  \\
    \midrule
MobileNetV2 (baseline) & &  94.1    & 76.3     & 83.8 & 74.0  & 46.3   \\
    \midrule
{\NAME} & {Weight Decay}
        &  95.0    & 77.8     & 89.0 & 84.4 & 49.1 \\
{\NAME} & {MixUp} &  94.1    & 77.0     & 85.2 & 79.6 & 47.4 \\
{\NAME} & {Arch}  &  94.5    & 76.8     & 84.1 & 76.4 & 46.3 \\
     \midrule
AutoHAS & {MixUp + Arch}
        &  94.4    & 77.4     & 84.8 & 78.2 & 47.3  \\
    \midrule
AutoHAS & {Weight Decay + MixUp}
        & \textbf{95.0 (\BLUE{+0.9})} & \textbf{78.4 (\BLUE{+2.1})} & \textbf{89.9 (\BLUE{+6.1})} & \textbf{84.4 (\BLUE{+10.4})} & \textbf{50.5 (\BLUE{+4.2})} \\
    \bottomrule
  \end{tabular}
  \caption{
  Classification top-1 accuracy (\%) of {\NAME} for different search space on five datasets.
  Weight decay and MixUp~\cite{zhang2018mixup} are for hyperparameters, and Arch is for architectures. Each experiment is repeated three times and the average accuracy is reported (standard deviation is about 0.2\%).
}
  \label{table:EXP-DATASET}
\end{table*}

\textbf{Searching settings.}
We call the hyperparameters that control the behavior of {\NAME} as meta hyperparameters -- the optimizer and learning rate for RL controller, the momentum ratio for RL baseline, and the warm-up ratio.
Warm-upping the REINFORCE algorithm indicates that we do not update the parameters of the controller at the beginning. In addition, when the search space includes architecture choices, we also uses the warm-up technique described in~\cite{bender2020can}.
For these meta hyperparameters, we use Adam, momentum as 0.95, warm-up ratio as 0.3. The meta learning rate is selected from \{0.01, 0.02, 0.05, 0.1\} according to the validation performance.
When the architecture choices are in the search space, we will use the absolute reward function~\citep{bender2020can} to constrain the FLOPs of the searched model to be the same as the baseline model.
For experiments on ImageNet and Places365, we use the batch size of 4096, search for 100 epochs, and use 4$\times$4 Cloud TPU V3 chips.
For experiments on other datasets, we use the batch size of 512, search for 15K steps, and use Cloud TPUv3-8 chips.\looseness=-1

\textbf{Training settings}.
Once we complete the searching procedure, we re-train the model using the AutoHAS-discovered hyperparameter and architecture.
For the components that are not searched for, we keep it the same as the baseline models.
For each experiment, we run three times and report the mean of the accuracy.

\subsection{Generalizability of {\NAME}}\label{sec:exp-gen}


To evaluate the generalization ability of {\NAME}, we have evaluated {\NAME} in different hyperparameter and architecture spaces for five datasets.
For simplicity, we choose the standard MobileNetV2~\citep{sandler2018mobilenetv2} as our baseline model\footnote{Baseline: the weight decay as 5e-6 and the mixup ratio as 0.}.
\Tabref{table:EXP-DATASET} shows the results.
We observe that AutoHAS achieves up to 10\% accuracy gain on the Flower dataset, suggesting that AutoHAS could be more useful for less optimized or new model/dataset scenarios.

\subsection{Efficiency of {\NAME}}\label{sec:exp-efficient}

We evaluate the efficiency of {\NAME} on ImageNet in \Figref{fig:autohas-vs-hpo}.
We still choose MobileNetV2~\citep{sandler2018mobilenetv2} as the baseline model. We search for the mixup ratio from [0, 0.2] and drop-path ratio from [0, 0.5] for each MBConv layer. We use the training schedule in~\citep{bender2020can}.
Results compared with four representative HPO methods are shown in \Figref{fig:autohas-vs-hpo}.
Multi-trial search methods -- Random Search~\citep{bergstra2012random} or Bayesian optimization~\citep{golovin2017google} -- must train and evaluate many candidates from \textit{scratch}, and thus are inefficient.
Even using 10$\times$ more time, they still cannot match the accuracy of {\NAME}.
This is because AutoHAS uses weight sharing and only requires negligible cost for every new sampled candidate. AutoHAS can traverse hundreds of more samples, within the same amount of time, than the traditional Random Search and Bayesian optimization.

Comparing AutoHAS to differentiable HPO methods, HGD~\citep{baydin2018hypergradient} can only search for the learning rate and the searched learning rate is much worse than the baseline.
IFT~\citep{lorraine2020optimizing} is an efficient gradient-based HPO method.
With the same search space, {\NAME} gets higher accuracy than IFT.


\begin{table}[ht!]
  \small\centering
  \setlength{\tabcolsep}{4pt}
  \begin{tabular}{l l l l l}
    \toprule
    Model  & Method & \#Params (M) & \#FLOPs (M) & Top-1 Accuracy (\%) \\
  \midrule
    \multirow{2}{*}{ResNet-50}
    & Human   & 25.6 & 4110  & 77.20 \\
    & AutoHAS & 25.6 & 4110  & \textbf{77.83} (\BLUE{+0.63}) \\
    \midrule
    \multirow{2}{*}{EfficientNet-B0}
    &   NAS   & 5.3 & 398 & 77.15 \\
    & AutoHAS & 5.2 & 418 & \textbf{77.92} (\BLUE{+0.77}) \\
    \bottomrule
  \end{tabular}
  \caption{
  AutoHAS improves popular models on the ImageNet dataset.
  }
  \label{table:imagenet-sota-model}
\end{table}

As another proof of {\NAME}'s efficiency, we follow MNasNet~\citep{tan2019mnasnet} and ProxylessNAS~\cite{cai2018proxylessnas} to design an architecture search space (i.e., kernel size \{3x3, 5x5\} and expansion ratio \{3, 6\} on top of MobileNetV2), and a joint search space with additional hyperparameter search options (i.e., mixup and dropout ratio).
We then compare {\NAME} performance on these two search spaces. With architecture-only search, AutoHAS achieves comparable results (e.g., 73.9\% accuracy @ 300M flops) as MNasNet/ProxylessNAS.
With the joint search, {\NAME} can further improve accuracy by 0.2\% with negligible additional cost.
Notably, {\NAME} has the the same level of computational costs as efficient NAS methods, but NAS methods are infeasible to optimize the hyperparameters.

\subsection{{\NAME} Can Scale to Large Datasets}\label{sec:exp-practical}

\begin{figure}[t!]

\begin{subfigure}{0.46\textwidth}
  \centering\includegraphics[width=\linewidth]{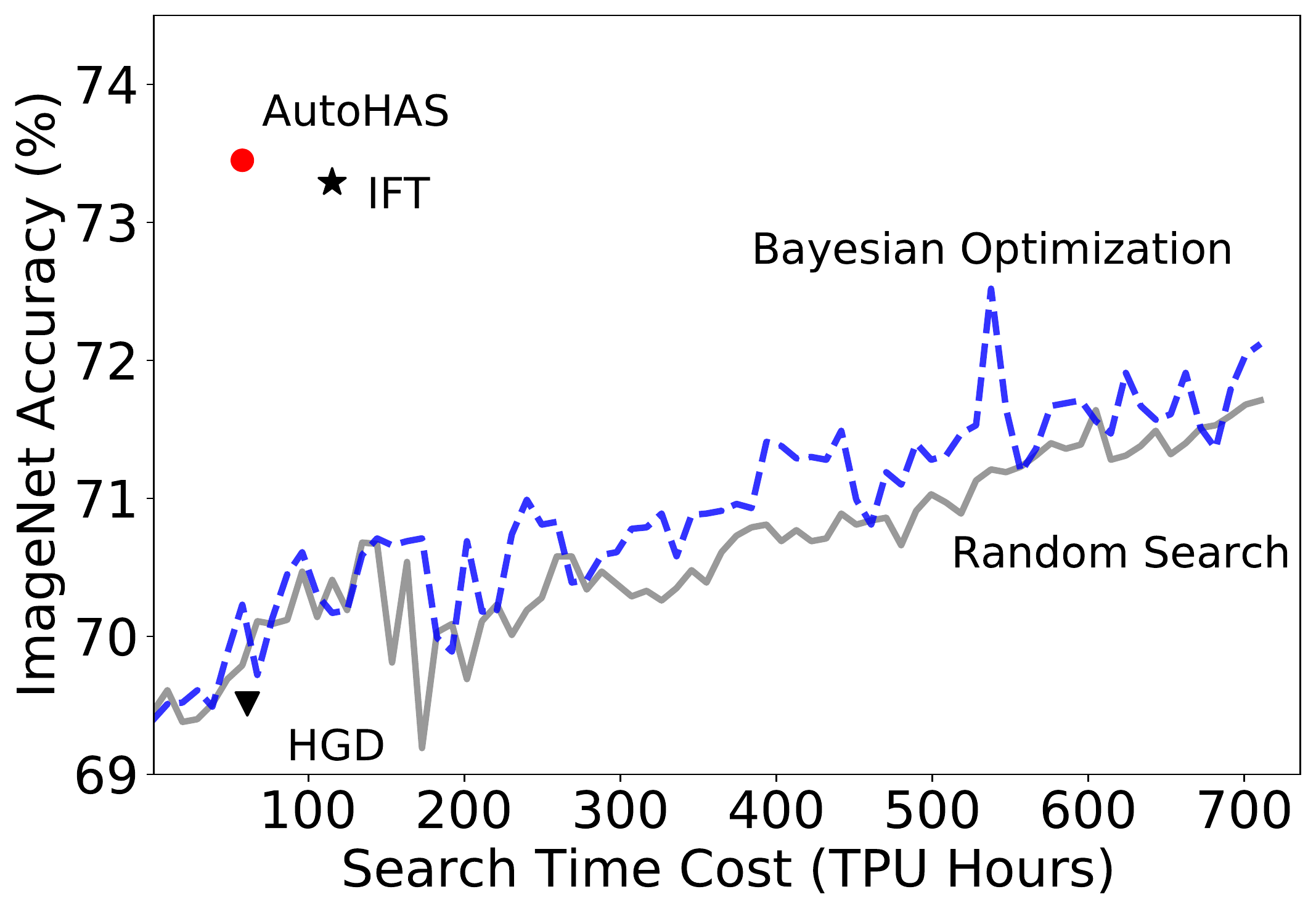}
  \caption{
Comparison between {\NAME} and previous HPO methods on ImageNet. AutoHAS achieves higher accuracy than HGD and IFT, and uses much less search time cost than others.
}\label{fig:autohas-vs-hpo}
\end{subfigure}
\hfill
\begin{subfigure}{0.46\textwidth}
  \centering\includegraphics[width=\linewidth]{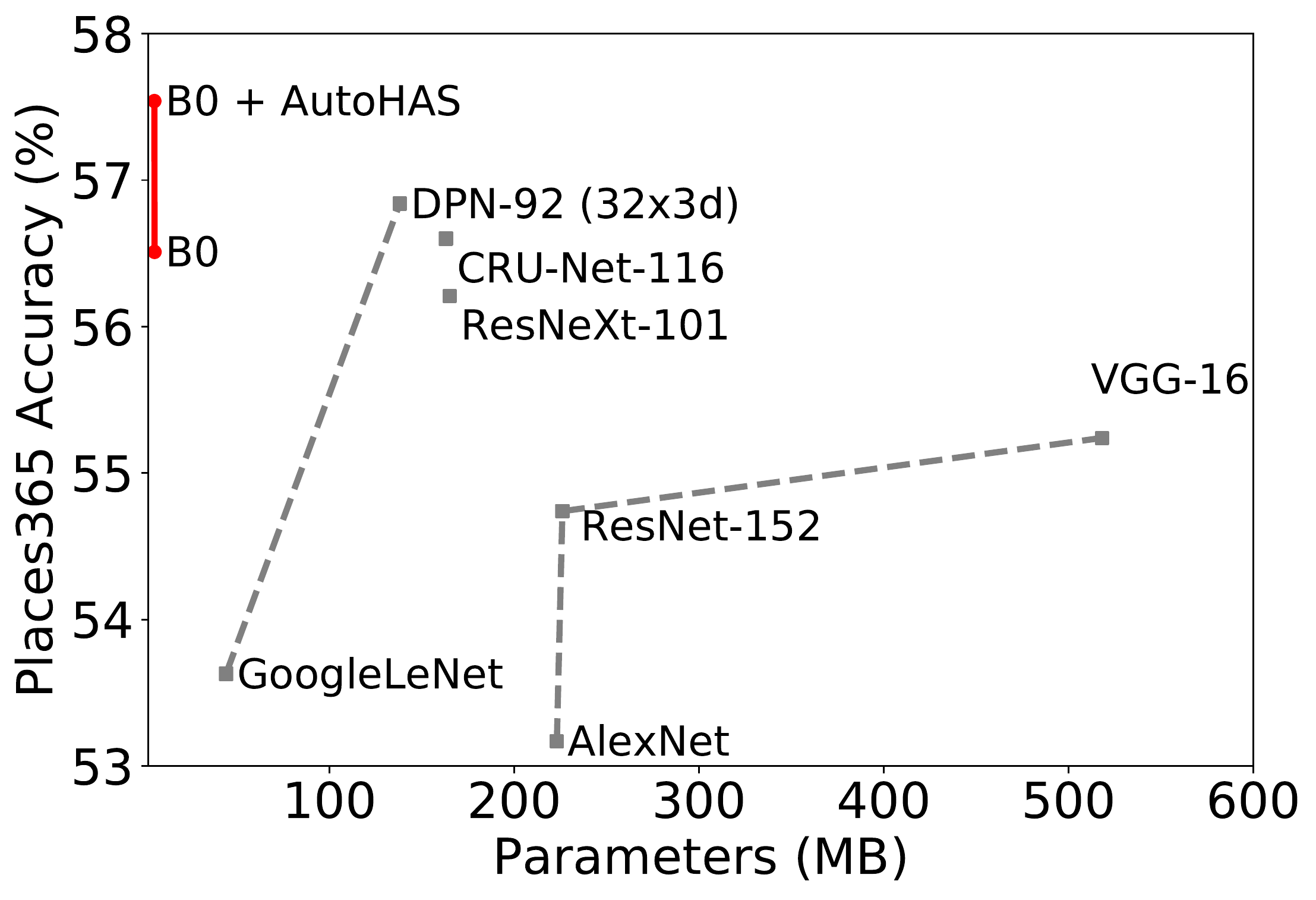}
  \caption{
{\NAME} improves accuracy by 1\% for EfficientNet-B0 on Places365.
}
\label{fig:autohas-places365}
\end{subfigure}
\caption{LEFT: Comparison between AutoHAS and other HPO methods.
RIGHT: Comparison between AutoHAS and other popular models.
}
\end{figure}

To investigate the effect of {\NAME} over the large-scale datasets,
we apply {\NAME} to two popular ImageNet models.
Firstly, we choose ResNet-50. The baseline strategy is to train it by 200 epochs, start the learning rate at 1.6 and decay it by 0.1 for every $\frac{1}{3}$ of the whole training procedure, use EMA with the decay rate of 0.9999, and apply SGD with the momentum of 0.9.
This can provide higher accuracy than the original paper. For reference, the reported top-1 accuracy is 76.15\% for ResNet-50 in TorchVision, whereas our baseline is 77.2\% accuracy.
Since previous methods usually do not tune the architecture of ResNet-50, we only use AutoHAS to search for its hyperparameters including the learning rate and the mixup ratio.
From \Tabref{table:imagenet-sota-model}, AutoHAS improves this ResNet baseline by 0.63\%.

\begin{table}[t!]
\scriptsize
  \centering
  \setlength{\tabcolsep}{3.4pt}
  \begin{tabular}{l c c c c c c}
    \toprule
    & \multirow{2}{*}{\#Params (MB)} & \multirow{2}{*}{\#FLOPs (M)} & \multirow{2}{*}{Accuracy (\%)} & \multicolumn{2}{c}{Search Cost}  \\
    &  & &  &  Memory (GB) & Time (TPU Hour) \\
    \midrule
 Baseline model & 1.5 & 35.9 & 50.96 &  1.0 & 44.8 \\
    \midrule
 AutoHAS (Differentiable Variant) & 1.5 & 36.1 & 52.17 &  6.1  & 92.8  \\
 AutoHAS (REINFORCE)       & 1.5 & 36.3 & 53.01 &  \textbf{1.8}  & \textbf{54.4}\\
 \bottomrule
  \end{tabular}
  \caption{
  AutoHAS Differentiable Search vs. AutoHAS REINFORCE Search -- Both are applied to the same baseline model with the same hyperparameter and architecture search space. Baseline model has no search cost, but we list its standalone training cost as a reference. Compared to the differentiable search, AutoHAS achieves slightly better accuracy with much less search memory cost.
  }
  \label{table:100-epochs-g-vs-r}
\end{table}

Secondly, we choose a NAS-searched model, EfficientNet-B0~\citep{tan2019efficientnet}. The baseline strategy is to train it by 600 epochs and use the same learning rate schedule as in the original paper.
As EfficientNet-B0 already tunes the kernel size and expansion ratio, we choose a different architecture space. Specifically, in each MBConv layer, we search for the number of groups for all the 1-by-1 convolution layer, the number of depth-wise convolution layer, whether to use a residual branch or not.
In terms of the hyperparameter space, we search for the per-layer drop-connect ratio, mixup ratio, and the learning rate.
We use {\NAME} to first search for the architecture and then for the hyperparameters.
From \Tabref{table:imagenet-sota-model}, we improves the strong EfficientNet-B0 baseline by 0.77\% ImageNet top-1 accuracy.

\textbf{{\NAME} improves SoTA Places365 models.}
Beside ImageNet, we have also evaluated AutoHAS on another popular dataset: Places365. Similarly, we apply AutoHAS to EfficientNet-B0 to search for better architectures and hyperparameters on this dataset.
\Figref{fig:autohas-places365} shows the results: Although EfficientNet-B0 is a strong baseline with significantly better parameter-accuracy trade-offs than other models, AutoHAS can still further improve its accuracy 1\% and obtain a new state-of-the-art accuracy on Places365.  Note that \TT{B0} and \TT{B0 + {\NAME}} only uses single crop evaluation, while other models use 10 crops.

\subsection{Ablation Studies and Visualization}\label{sec:exp-abla}

\begin{figure}[ht!]
  \begin{center}
    \includegraphics[width=\linewidth]{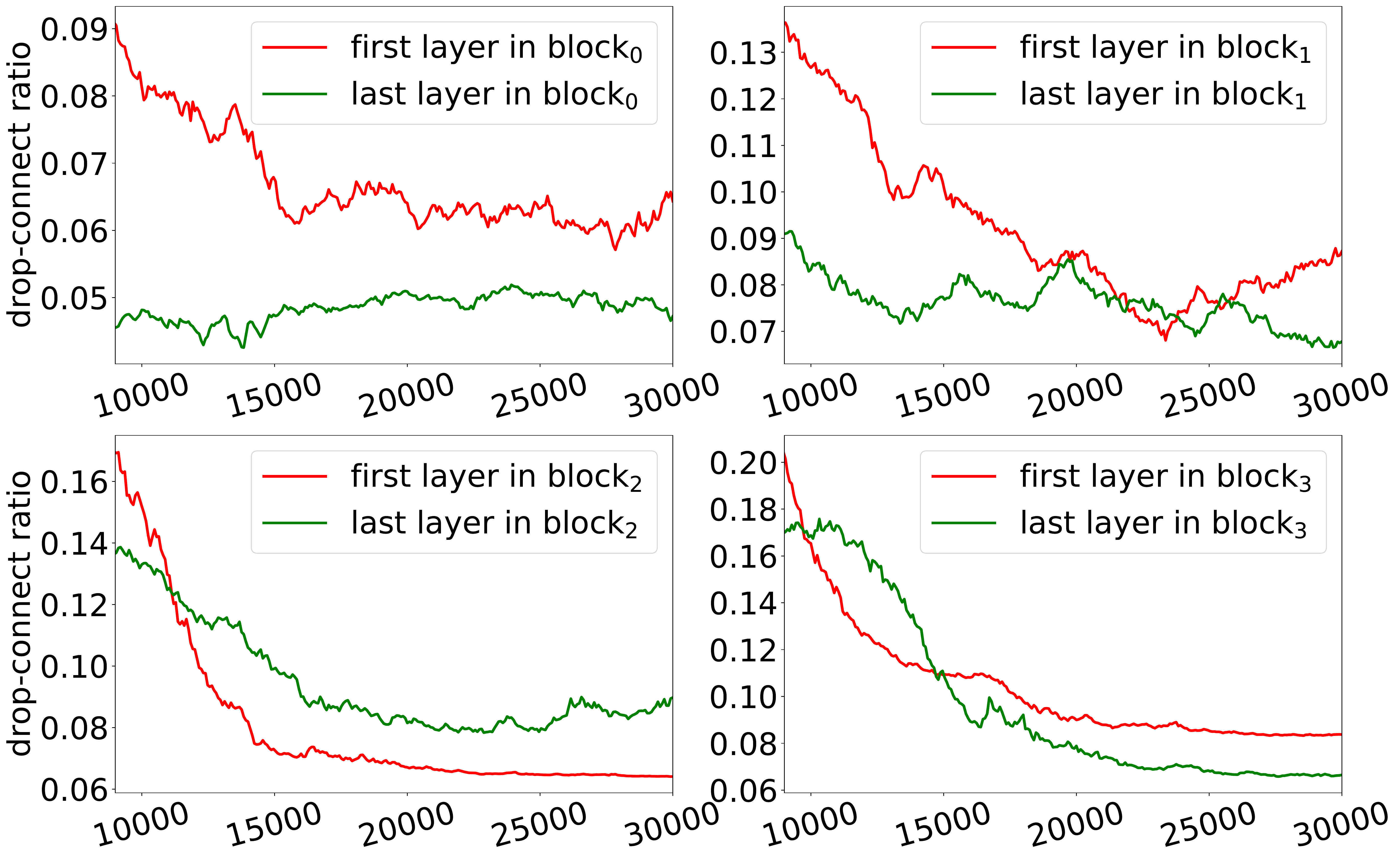}
  \end{center}
\caption{
We visualize the AutoHAS-discovered drop-connect ratio for some layers in EfficientNet-B0.
The x-axis and y-axis indicate the search step and drop-connect ratio, respectively.
}
\label{fig:autohas-vis}
\end{figure}

\textbf{Why choose RL instead of a differentiable strategy?}
Differentiable search methods have been extensively studied for its simplicity in many previous literature~\citep{liu2019darts,dong2019search,wan2020fbnetv2,xie2019snas}, but these methods usually require much higher memory cost in order to train the entire super model.
In our AutoHAS framework, we employ a simple reinforcement learning algorithm -- REINFORCE~\citep{williams1992simple} -- to optimize the controller: instead of training the whole super model, we only train a subset of the super model and therefore significantly reduce the training memory cost.
Notably, the REINFORCE could also be replaced by a differentiable-based algorithm with the supervision of validation loss.
We investigate the difference between differentiable and REINFORCE search in \Tabref{table:100-epochs-g-vs-r}.
Not surprisingly, differentiable search requires much higher memory cost (6.1x more than baseline) as it needs to maintain the feature or gradient tensors for all the super model, whereas our REINFORMCE-based AutoHAS is much more memory efficient: reducing the memory cost by 70\% than the differentiable approach. Empirically, we observe they achieve similar accuracy gains in this case, but AutoHAS enables us to search for much larger models such as EfficientNet-B0 and ResNet-50 as shown in \Tabref{table:imagenet-sota-model}.

\textbf{Visualization}.
We show the intermediate search results of drop-connect ratios in \Figref{fig:autohas-vis}.
Human experts have prior knowledge that drop-connect is crucial to the performance.
However, it is prohibitive to manually tune a proper ratio for each of those 10+ drop-connect layers in EfficientNet-B0.
Our {\NAME} is suitable for such a scenario, it can discover suitable ratios for many drop-connect layers in a single trial.

\section{Conclusion \& Future Work}\label{sec:conclusion}

In this paper, we proposed an automated and unified framework {\NAME}, which can efficiently search for hyperparameters and architectures.
{\NAME} provides a novel perspective of AutoML algorithms by generalizing the weight sharing technique from architectures to hyperparameters.
{\NAME} integrates several techniques to be memory friendly, efficient, generalized, and scalable.
Experimentally, {\NAME} improves the baseline models on \textit{eight} datasets over different kinds of search spaces.
For the highly-optimized ResNet/EfficientNet, it improves ImageNet top-1 accuracy by 0.8\%; for other less-optimized scenarios (e.g., Oxford Flower), it improves the accuracy by 11.4\%.

In the future, we would evaluate AutoHAS on NAS or HAS benchmarks.
In addition, we would comprehensively investigate AutoHAS's performance by integrating different searching algorithms in AutoHAS.

\end{document}